\documentclass{article}
\usepackage{ijcai22}

\usepackage{times}

\usepackage{soul}
\usepackage{url}
\usepackage[hidelinks]{hyperref}
\usepackage[utf8]{inputenc}
\usepackage[small]{caption}
\usepackage{graphicx}
\usepackage{amsmath}
\usepackage{booktabs}
\usepackage{verbatim}
\title{Improving Model Understanding and Trust with Counterfactual Explanations of Model Confidence}

\author{
	Thao Le\footnote{Contact Author}\and
	Tim Miller\and
	Ronal Singh\And
	Liz Sonenberg\\
	\affiliations
	The University of Melbourne
	\emails
	thaol4@student.unimelb.edu.au,
	\{tmiller, rr.singh, l.sonenberg\}@unimelb.edu.au
}

\DeclareMathOperator*{\argmin}{arg\,min}

\usepackage{subcaption}
\usepackage{bm}
\usepackage{multirow}
\usepackage{algorithm}
\usepackage{algorithmic}
\usepackage{float}
\usepackage{placeins}
\usepackage{tikz}
\newcommand*{\checkmark}[1][]{\tikz[x=1em, y=1em]\fill[#1] (0,.35) -- (.25,0) -- (1,.7) -- (.25,.15) -- cycle;}

\begin{document}

\maketitle

\begin{abstract}
In this paper, we show that counterfactual explanations of confidence scores help users better understand and better trust an AI model's prediction in human-subject studies. Showing confidence scores in human-agent interaction systems can help build trust between humans and AI systems. However, most existing research only used the confidence score as a form of communication, and we still lack ways to explain why the algorithm is confident. This paper also presents two methods for understanding model confidence using counterfactual explanation: (1) based on counterfactual examples; and (2) based on visualisation of the counterfactual space.
\end{abstract}

\section{Introduction}
Explaining why the AI model gives a certain prediction can promote trust and understanding for users, especially for non-expert users. While recent research~\cite{Zhang20effect,Wang21uncertainty} has used confidence (or uncertainty) measures as a way to improve AI model understanding and trust, the area of explaining why the AI model is confident (or not confident) in its prediction is still underexplored~\cite{TomsettPBCCSPK20}.

In Machine Learning (ML), the \textit{confidence score} indicates the chances that the ML model's prediction is correct. In other words, it shows how \textit{certain} the model is in its prediction, which can be defined as the predicted probability for the best outcome~\cite{Zhang20effect}. Another way to define the \textit{confidence score} is based on the uncertainty measures, which can be calculated using entropy ~\cite{Bhatt20uncertainty} or using \textit{uncertainty sampling}~\cite{LewisG94},~\cite[p93]{monarch2021human}.

In this paper, we will complement prior research by applying a counterfactual (CF) explanation method to generate explanations of the confidence of an output prediction. It is increasingly accepted that explainability techniques should be built on research in philosophy, psychology and cognitive science~\cite{Miller19social,Byrne19} and that the evaluation process of explanation should involve human-subject studies~\cite{Miller17Beware,ForsterHKK21,KennyFQK21,WaaNCN21}. We therefore evaluate our explanation to know whether counterfactual explanation can improve \textit{understanding} and \textit{trust} in two user studies. We present the CF explanation using two designs: (1) providing counterfactual instances in a table; and (2) visualising the counterfactual space for each feature and its effect on model confidence. We also investigate whether the different designs result in a significant difference in user \textit{satisfaction} with the explanation.

In summary, our contributions are:
\begin{itemize}
    \item We formalise two approaches for counterfactual explanation of confidence score: one using counterfactual examples and one visualising the counterfactual space.  
    \item Through two user studies we demonstrate that showing counterfactual explanations of confidence scores can help users better understand and trust the AI model.
\end{itemize}

\section{Background}

In this section, we review related works on counterfactual explanations and confidence (or uncertainty) measures.

\subsection{Counterfactual Explanations}
Counterfactual explanation is described as the possible smallest changes in input values in order to change the model prediction to a desired output~\cite{wachter2017counterfactual}. It has been increasingly used in explainable AI (XAI) to facilitate human interaction with the AI model~\cite{Miller19social,Miller2021-ql,Byrne19,ForsterHKK21}. Counterfactual explanations are expressed in the following form: ``You were denied a loan because your annual income was \$30,000. If your income had been \$45,000, you would have been offered a loan". To generate counterfactuals, \cite{wachter2017counterfactual} suggest finding solutions of the following loss function.
\begin{equation}
\argmin_{x'}\max_{\lambda} \lambda(f(x') - y')^2 + d(x,x')
\label{loss-wachter}	
\end{equation}
where $x'$ is the counterfactual solution; $(f(x') - y')^2$ presents the distance between the model's prediction output of counterfactual input $x'$ and the desired counterfactual output $y'$; $d(x,x')$ is the distance between the original input and the counterfactual input; and $\lambda$ is a weight parameter. A high $\lambda$ means we prefer to find counterfactual point $x'$ that gives output $f(x')$ close to the desired output $y'$, a low $\lambda$ means we aim to find counterfactual input $x'$ that is close to the original input $x$ even when the counterfactual output $f(x')$ can be far away from the desired output $y'$. This loss function can be solved by iteratively increasing $\lambda$ until a close solution $x'$ is found. In this model, $f(x)$ would be the output, such as a denied loan, and $y'$ would be the desired output -- the loan is granted. The counterfactual $x'$ would be the properties of a similar customer that would have received the loan.

\cite{Russell19} proposes a search algorithm to generate counterfactual explanations based on mixed-integer programming, assumed where input variables can be continuous or discrete values. They defined a set of linear integer constraints, which is called \textit{mixed polytope}. These constraints can be given to Gurobi Optimization~\cite{gurobi_2022} and then an optimal solution is generated. They find the counterfactual point $x'$ by solving this function.

\begin{equation}
    \argmin_{x'} ||\hat{x} - x'||_{1,w}
    \label{eq:search-counterfactual}
\end{equation}

where $\hat{x}$ is the mixed encoding of $x$; $x'$ lies on the mixed polytope; $||.||_{1,w}$ is a weighted $l_1$ norm with weight $w$ is defined as the inverse median absolute deviation (MAD)~\cite{wachter2017counterfactual}.

\cite{Antoran21explainUncertainty} propose Counterfactual Latent Uncertainty Explanations (CLUE), to identify which input features are responsible for the model's uncertainty. Their model for finding counterfactual examples is similar to ours, however we go further by considering ways to visualise the counterfactual space, and run a more comprehensive user study to measure understanding, satisfaction, and trust.

There are many other approaches to solving counterfactuals for tabular~\cite{MothilalST20,KeaneS20}, image~\cite{GoyalWEBPL19,ChangCGD19,dhurandhar-contrastive}, text~\cite{JacoviSRECG21,RiveiroT21} and time series data~\cite{DelaneyGK21}. None of these are for explaining model confidence, however, the underlying algorithms could be modified to search over the model confidence instead of the model output.

\subsection{Confidence (Uncertainty) measures}

A confidence score measures how confident a ML model is in its prediction; or inversely, how uncertain it is. A common method of measuring uncertainty is to use the prediction probability~\cite{Delaney21,Bhatt20uncertainty}. Specifically, \textit{uncertainty sampling }~\cite{LewisG94} is an approach that queries unlabelled instance $x$ with maximum uncertainty to get human feedback. There are four types of uncertainty sampling~\cite[p70]{monarch2021human}: \textit{Least confidence}, \textit{Margin of confidence}, \textit{Ratio of confidence} and \textit{Entropy}. 

\cite{Zhang20effect} demonstrate that communicating confidence scores can support trust calibration for end users. \cite{Wang21uncertainty} also argue that showing feature attribution uncertainty helps improve model understanding and trust. The basis of the feature attribution is that it is a technique to explain the AI model by measuring how input values affect the output prediction~\cite{Lundberg17Unified,Ribeiro-explain-any-classifier}. 

\cite{WaaSDN20} propose a framework called \textit{Interpretable Confidence Measures (ICM)} which provides predictable and explainable confidence measures based on case-based reasoning~\cite{Atkeson97}. Case-based reasoning provides prediction based on similar past cases of the current instance. This approach however did not address counterfactual explanations of model confidence.
 
\section{Formalising Counterfactual (CF) Explanation of Confidence}

In this section, we describe two methods for CF explanation: one based on counterfactual examples~\cite{Antoran21explainUncertainty} and one based on counterfactual visualisation as in Figure~\ref{fig:graph-explanation}.

\subsection{Generating Counterfactual Explanation of Confidence}

In this section, we show how to generate counterfactual explanations of the confidence score in data where input variables can take either discrete or continuous values. For example, when the AI model predicts that an employee will leave the company with confidence of $70\%$, a person may ask: \emph{Why is the model 70\% confident instead of 40\% confident or less?}. We aim to generate counterfactual inputs that bring the confidence score to 40\% or lower. An example of counterfactual explanation in this case is: \textit{``One way you could have got a confidence score of 40\% instead is if Daily Rate had taken the value 400 rather than 300"}. Therefore, from this counterfactual explanation, we know that we need to increase the employee's daily rate to lower the confidence of them resigning from the company.

We now propose our approach to generate counterfactual explanations of confidence scores. We follow \cite{Russell19} in proposing the algorithm to search for counterfactual points of output confidence by seeking the counterfactual point $x'$ in the \textit{mixed polytope} based on Equation~\ref{eq:search-counterfactual}. However, we add three new constraints to ensure that we find the counterfactual points that change the confidence score but do not change the output prediction. Three constraints are described as follows.

Given factual confidence score $T$ (can be specified by the `user'), the counterfactual explanation of confidence $U(x')$  generated by data point $x'$ is found by one of the following two constraints:
\begin{align}
	U(x') > T \label{eq:contrastive-greater} \\
	U(x') < T \label{eq:contrastive-lesser}
\end{align}
where $T$ is the factual confidence score; $U(x')$ is the counterfactual confidence score. We apply Equation~\ref{eq:contrastive-greater} when we want to find counterfactual $x'$ that increases the confidence score, and Equation~\ref{eq:contrastive-lesser} for a counterfactual $x'$ that decreases the confidence score. We call Equation~\ref{eq:contrastive-greater} and~\ref{eq:contrastive-lesser} \textit{Constraint 1}.

Note that $x$ (input instance) and $x'$ will give the same output prediction as class $k$ but different confidence scores $U(x)$ and $U(x')$. To meet this condition, $P(x)$ and $P(x')$ must be in the same space according to the decision boundary. We call this \textit{Constraint 2}, and define it as:

\begin{equation}
    \begin{cases}
        P(y=k|x') < D & \text{if } P(y=k|x) < D \\
        P(y=k|x') >= D & \text{if } P(y=k|x) >= D
    \end{cases}
\end{equation}
where $D$ is the decision boundary that classifies the class.

The final constraint that must hold is $x \neq x'$ in order to search for the counterfactual point $x'$ that is not equal to the original point $x$ (\textit{Constraint 3}). This constraint is important in two cases
\begin{itemize}
    \item setting $U(x) > T$ and seeking $x'$ based on $U(x') > T$.
    \item setting $U(x) < T$ and seeking $x'$ based on $U(x') < T$
\end{itemize}

In our experiments, we use logistic regression to calculate the probability of a class, so $P(x) = \frac{1}{1+e^{-y}}$ where $y=wx$ is a linear function of point $x$. We choose \textit{margin of confidence}, which is the difference between the first and the second highest probabilities~\cite[p93]{monarch2021human} as the formula of confidence score $U(x)$. The higher the difference between two class probabilities, the more confident the prediction is in the highest probability class. In binary classification, the score is minimum when probability $P(x) = 0.5$, which is the decision boundary of binary classification. $U(x)$ is then calculated as follows:
\begin{displaymath}
U(x) = |P(x) - (1 - P(x))| = |2P(x) - 1|
\end{displaymath}

Then,
\begin{equation}
	U(x) = 
	\begin{cases}
		2P(x) - 1 &\text{if } P(x) \geq 0.5 \\
		1 - 2P(x)  &\text{otherwise }\\
	\end{cases}
\label{eq:confidence-score}
\end{equation}

\subsection{Examples of counterfactual explanation of confidence score}

\begin{table}[ht]
	\centering \small
	\setlength{\tabcolsep}{1pt} % Default value: 6pt
	\renewcommand{\arraystretch}{1} % Default value: 1
  \begin{tabular}{lccc}
    \toprule
  \textbf{Attribute} & \textbf{Alternative 1}  & \textbf{Alternative 2} & \textbf{Original Value}\\
  \midrule
\textbf{Marital Status} & \textbf{Married} & \textbf{Never}  & \textbf{Divorced/}\\
& & \textbf{Married} & \textbf{Widowed}\\
%\midrule
Years of Education&	-&	- & 10\\
%\midrule
Occupation&	-& - & Service\\
%\midrule
Age & - & - & 34\\
%\midrule
Any capital gains&	-&	- & No\\
%\midrule
Working hours &-&-& 37\\
per week  & & & \\
%\midrule
Education &	-  & - & Professional or \\
& & & Associate Degree \\
\midrule
\textbf{Confidence score} &	\bm{$43.6\%$} & \bm{$94.4\%$} &	\bm{$91.5\%$}\\
\midrule
\textbf{AI prediction}	& \multicolumn{3}{c}{\textbf{Lower than \$50,000}} \\
  \bottomrule
\end{tabular}
\caption{Example of counterfactual explanations as example instances. In alternative columns, notation (-) means the value is unchanged from the original value, we only highlight the values that changed.}
\label{tab:table-explanation}
\end{table}

Given the original instance input shown in column \textit{Original Value} in Table~\ref{tab:table-explanation}, the AI model predicts that this person has an income of \textit{Lower than $\$50,000$} with a confidence score of $91.5\%$. We choose a factual confidence score $T = 0.5$ and search for $x'$ where $U(x') < T$. An example of counterfactual explanation generated using our method is:
\textit{``One way you could have got a confidence score of less than 0.5 (0.44) instead is if Marital Status had taken value Married rather than Divorced/Widowed."}

In the first method, we use a table with an example provided in Table~\ref{tab:table-explanation}. In this table explanation, we show the details of a person in column \textit{Original Value} and the prediction that their income is lower than $\$50,000$. When we change the value of feature \textit{Marital Status} as in columns \textit{Alternative 1} and \textit{Alternative 2}, the confidence score changes but the prediction is still lower than $\$50,000$. From this table, we can find the correlation between the \textit{Marital Status} and the confidence score such that the marital status \textit{Never Married} gives the prediction with the highest confidence score among all three marital statuses.

\subsection{Counterfactual Visualisation}

\begin{figure}[!htbp]
	\centering
	\includegraphics[width=0.8\linewidth]{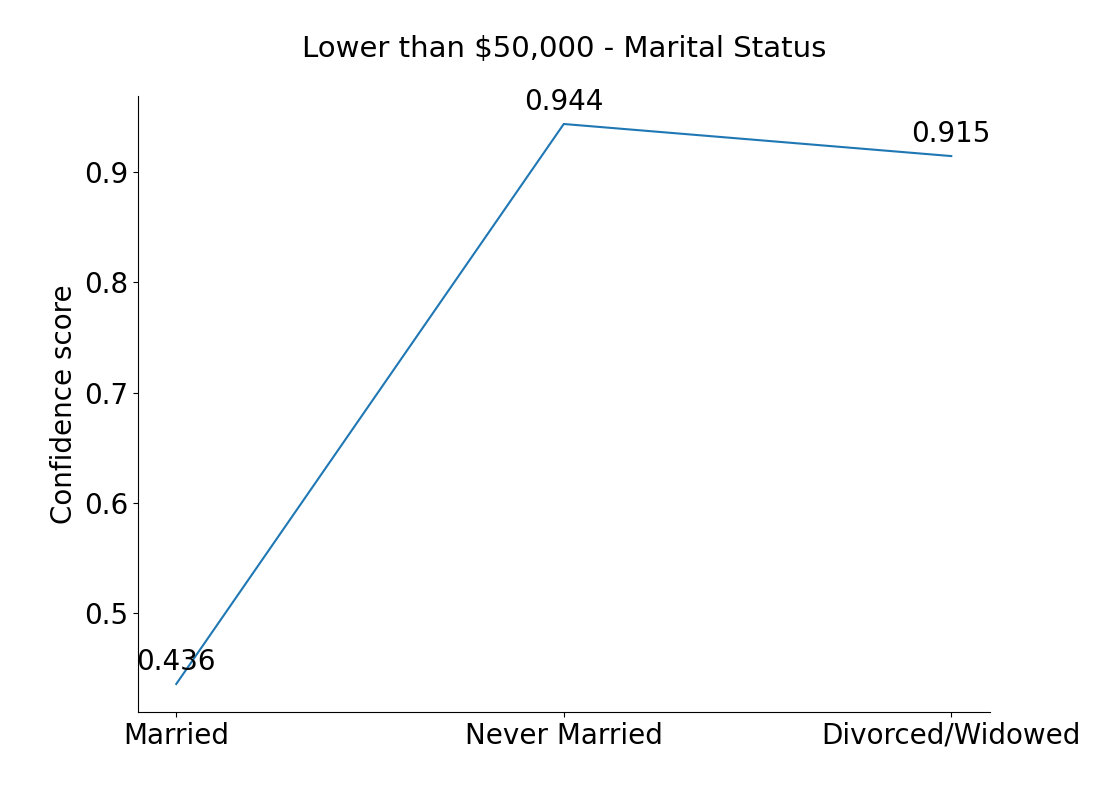}
	\caption{Example of counterfactual visualisation}
	\label{fig:graph-explanation}
\end{figure}

In this section, we propose a method for visualising the counterfactual space of a model and how this affects the model's confidence. The idea is to show how varying a single feature affects the model's confidence. For example, Figure~\ref{fig:graph-explanation} shows the visualisation based on Table~\ref{tab:table-explanation} in the income prediction task. Here we can see the prediction reaches maximum confidence score at \textit{Never Married} status. The title of this graph shows the output prediction \textit{Lower than $\$50,000$} and the feature name \textit{Marital Status} which we used to change the values.

This graph visualisation technique is based on the idea of \textbf{Individual Conditional Expectation (ICE)}~\cite{Goldstein15}. ICE is often used to show the effect of a feature value on the predicted probability of an instance. In our study, we show how changing a feature value can change the \textit{confidence score} instead of the predicted probability. There are two types of variables in the dataset: (1) discrete variable, and (2) continuous variable. So we define the ICE for confidence score of a single feature $x_i$ of instance $x$ such that:

\begin{equation}
    F(x_i) = U(x_i) = |2P(x_i) - 1|
\end{equation}
for all $x_i$, where:
\begin{itemize}
    \item $x_i \in D$ if $x_i$ is a discrete value and $D$ is the discrete set
    \item $x_i \in [c_{\min}, c_{\min}+t, \ldots, c_{\max}]$ if $x_i$ is a continuous value; $c_{\min}$ and $c_{\max}$ are the minimum and maximum values of a continuous range and $t$ is a fixed increment.
\end{itemize}

For the graph explanation, it can only change one feature at a time whereas counterfactual examples presented in tables can change many features. However, the graph explanation might be easier to identify the maximum (or minimum) value of discrete values and the trend of continuous values.

\section{Human-Subject Experiments}

\begin{table*}[!t]
	\centering \small
	\renewcommand{\arraystretch}{1.2} % Default value: 1
	\begin{tabular}{c p{0.2\textwidth}  p{0.25\textwidth}  p{0.25\textwidth}}
		\toprule
		& \textbf{Control} & \textbf{Treatment - Table Explanation} & \textbf{Treatment - Graph Explanation}\\
		\midrule
		Phase 1 & \multicolumn{3}{c}{
			Participants are given plain language statement, consent form and demographic questions (age, gender)
		}\\
		%\midrule
		\begin{comment}
		Phase 2 & \multicolumn{3}{c}{Participants are provided with}\\
		& \begin{itemize}
			\item Input instances
			\item AI model's prediction class
		\end{itemize}
		& 
		\begin{itemize}
			\item Counterfactual examples presented in a table
			\item Input instances
			\item AI model's prediction class
		\end{itemize}
		&
		\begin{itemize}
			\item Counterfactual visualisation presented in a graph	    
			\item Input instances
			\item AI model's prediction class
		\end{itemize}
		\\
		& \multicolumn{3}{c}{
			Participants will choose the instance which the AI model will predict with the highest confidence score
		}\\
		\end{comment}
		Phase 2 & \multicolumn{3}{c}{Participants are provided with}\\
		 & Input instances & Input instances & Input instances\\
		& AI model's prediction class & AI model's prediction class & AI model's prediction class\\
		& & Counterfactual examples presented in a table & Counterfactual visualisation presented in a graph\\
		%\midrule
		Phase 3 & \multicolumn{1}{c}{Nothing} & \multicolumn{2}{c}{10-point Likert \textit{Explanation Satisfaction Scale}} \\
		%\midrule
		Phase 4 & \multicolumn{3}{c}{10-point Likert \textit{Trust Scale}} \\
		\bottomrule
	\end{tabular}
	\caption{Summary of participants' tasks in our three experimental conditions}
	\label{tab:task-groups}
\end{table*}

Our user experiments test the following hypotheses:
\begin{itemize}
\item \textbf{Conjecture 1}: Counterfactual explanations of confidence score help user build mental models leading to a \textbf{better understanding} of why the AI model is confident in its prediction.
    \begin{itemize}
        \item \textbf{Hypothesis 1a (H1a)}: Table counterfactual explanations help users better \textbf{understand} the AI model than when they are not given explanations.
        \item \textbf{Hypothesis 1b (H1b)}: Graph counterfactual explanations help users better \textbf{understand} the AI model than when they are not given explanations.
    \end{itemize}
\item \textbf{Conjecture 2}: Counterfactual explanations of confidence score improve user \textbf{trust} in the AI model more than when they are not given explanations.
    \begin{itemize}
        \item \textbf{Hypothesis 2a (H2a)}: Table counterfactual explanations help users better \textbf{trust} the AI model than when they are not given explanations.
        \item \textbf{Hypothesis 2b (H2b)}: Graph counterfactual explanations help users better \textbf{trust} the AI model than when they are not given explanations.
    \end{itemize}
\end{itemize}
We then evaluate the difference between counterfactual explanations presented in a table and counterfactual in a graph based on the following hypotheses.
\begin{itemize}
    \item \textbf{Hypothesis 3a (H3a)}: Graph counterfactual explanations help users better \textbf{understand} the AI model than table counterfactual explanations.
    \item \textbf{Hypothesis 3b (H3b)}: Graph counterfactual explanations help users better \textbf{trust} the AI model than table counterfactual explanations.
    \item \textbf{Hypothesis 3c (H3c)}: Graph counterfactual explanations improve user \textbf{satisfaction} in explanation more than showing table counterfactual explanations.
\end{itemize}

To evaluate H1a, H1b and H3a, we use \textit{task prediction}~\cite[p11]{Hoffman-metrics-xai}. In task prediction, participants are given some instances and their task is to decide for which instance the AI model will predict a higher confidence score. Thus, \textit{task prediction} helps evaluate the user's mental model about their understanding in model confidence. To test H2a, H2b and H3b, we use 10-point Likert \textit{Trust Scale} from~\cite[p49]{Hoffman-metrics-xai}. Finally, we use 10-point Likert \textit{Explanation Satisfaction Scale} from~\cite[p39]{Hoffman-metrics-xai} to evaluate H3c.

\subsection{Experimental Design}

\begin{table}
	\centering \small
	\setlength{\tabcolsep}{1pt} % Default value: 6pt
	\renewcommand{\arraystretch}{1} % Default value: 1
	\begin{tabular}{lccc}
		\toprule
		\textbf{Attribute} & \textbf{Employee 1} & \textbf{Employee 2} & \textbf{Employee 3} \\
		\midrule
		Marital Status & \textbf{Divorced/} & \textbf{Married} & \textbf{Never} \\
	    %\midrule
	    & \textbf{Widowed} & & \textbf{Married} \\
		Years of Education & 15 & 15 & 15\\
		%\midrule
		Occupation & Service & Service & Service\\ 
		%\midrule
		Age & 25 & 25 & 25\\
		%\midrule
		Any capital gains & No & No & No\\
		%\midrule
		Working hours & 30 & 30 & 30\\
		per week & & \\
		%\midrule
		Education & Bachelors &  Bachelors & Bachelors\\
		\midrule
		\textbf{AI model prediction} & \multicolumn{3}{c}{\textbf{Lower than }\bm{$\$50,000$}} \\
		\bottomrule
	\end{tabular}
	\caption{Example input instances provided in the question. The question is: ``For which employee the AI model predicts with the highest confidence score?"}
	\label{tab:example-question}
\end{table}

\paragraph{Dataset} We ran the experiment on two different domains from two different datasets, which are \textit{income prediction domain} and \textit{HR domain}. The data used for the income prediction task is the Adult Dataset published in UCI Machine Learning Repository~\cite{Dua-2019} that includes 32561 instances and 14 features. This dataset classifies a person's income into two classes (below or above \$50K) based on personal information such as marital status, age, and education. In the second domain, we use the IBM HR Analytics Employee Attrition Performance data published in Kaggle~\cite{pavansubhash_2017} which includes 1470 instances and 34 features. This dataset classifies employee attrition as yes or no based on some demographic information (job role, daily rate, age, etc.). We selected the seven most important features for both datasets by applying the Gradient Boosting Classification model over all data.

\paragraph{Procedure} Before conducting the experiments, we received ethics approval from our institution. We recruited participants on Amazon Mechanical Turk (Amazon MTurk), a popular crowd-sourcing platform for human-subject experiments~\cite{buhrmester2016amazon}. The experiment was designed as a Qualtrics survey~\footnote{\url{https://www.qualtrics.com/}} and participants can navigate to the survey through the Amazon MTurk interface. We allowed participants 30 minutes to finish the experiment and paid each participant a minimum of USD \$7 for their time. They also had a chance of winning a maximum bonus of USD \$2 depending on their final score.

In each domain, we use a between-subject design such that participants were randomly assigned into one of three groups (\textit{Control}, \textit{Treatment with Table Explanation}, \textit{Treatment with Graph Explanation}). For each group, there are four phases that are described in Table~\ref{tab:task-groups}. First, participants were given a plain language statement that described their task and a consent form. After they gave consent to do the study, they were asked demographic questions (age range, gender). Then each participant was randomly allocated to one of the three groups. The difference between the control group and the treatment group is that in the control group, participants were not given any counterfactual explanations. In the task prediction (phase 2), they were only shown input values along with the AI model prediction class as in Table~\ref{tab:example-question}. 
In the treatment group, participants were provided with counterfactual explanations presented either in a table or graph. Two examples of the counterfactual explanations are shown in Table~\ref{tab:table-explanation} and Figure ~\ref{fig:graph-explanation}. The participants each received the same 10 tasks. For each task, they were asked to select an input instance out of \textit{3 instances} that the AI model would predict with the highest confidence score. We scored each participant using: 1 for a correct answer, -2 for a wrong answer and 0 for selecting ``I don't have enough information to decide". To imitate high-stake domains, the loss for a wrong choice is higher than the reward for a correct choice~\cite[p2433]{Bansal19}. They are also asked to briefly explain why they choose that option in a text box. The final compensation was calculated based on the final score --- a score of $0$ or less than $0$ received \$7 USD and no bonus. A score greater than $0$ received a bonus of \$0.2 for each additional score. In phase 3, participants assigned to treatment groups evaluated the explanations based on a continuous 10-point Likert \textit{Explanation Satisfaction Scale}~\cite[p39]{Hoffman-metrics-xai}. Finally, all participants evaluated their trust in the AI model based on a continuous 10-point Likert \textit{Trust Scale}~\cite[p49]{Hoffman-metrics-xai}.

In summary, there are three independent variables in our experiment: (1) Control group with no explanations (\textit{C}), (2) Treatment group using table explanations (\textit{TT}) and (3) Treatment group using graph explanations (\textit{TG}).

\paragraph{Participants} We recruited a total of 180 participants for two domains, that is 90 participants for each domain from Amazon MTurk. Then 90 participants were evenly randomly allocated into three groups (30 participants in each group). All participants were from the United States. We only recruited Masters workers, who achieved a high degree of success in their performance across a large number of Requesters~\footnote{\url{https://www.mturk.com/worker/help}}. For the \textit{income prediction domain}, 41 participants were women, 1 was self-specified as non-binary, 48 were men. Between them, 4 participants were between Age 18 and 29, 34 between Age 30 and 39, 27 between Age 40 and 49, 25 over Age 50. For the \textit{HR domain}, 43 participants were women, 47 were men. Age wise, 4 participant was between Age 18 and 19, 37 between Age 30 and 39, 26 between Age 40 and 49, 23 over Age 50.

We performed \textit{power analysis} for two independent sample t-test to determine the needed sample sizes. We calculate the Cohen's d between control and treatment group and obtain the effect size of $0.7$ and $0.67$ in income and HR domain. Using power of $0.8$ and significant alpha of $0.05$, we get sample sizes of $26$ and $29$ in the two domains. Thus, we determine the sample size needed for a group is $30$ and the total number of samples needed is $90$ for one domain.

\subsection{Results: Domain 1 - Income Prediction}

\begin{comment}
\begin{figure}[ht]
	\begin{subfigure}[b]{0.49\columnwidth}
		\centering
		\includegraphics[width=\linewidth]{fig/income_score_by_group}
		\caption{Domain 1 (Income)}
		\label{fig:income-final-score}
	\end{subfigure}
 	\begin{subfigure}[b]{0.49\columnwidth}
		\centering
		\includegraphics[width=\linewidth]{fig/hr_score_by_group} %height=3cm
		\caption{Domain 2 (HR)}
		\label{fig:hr-final-score}
	\end{subfigure}
	\caption{Final score in two domains}
\end{figure}
\end{comment}

\begin{figure}[ht]
    \begin{subfigure}[b]{0.33\columnwidth}
		\centering
		\includegraphics[width=\linewidth]{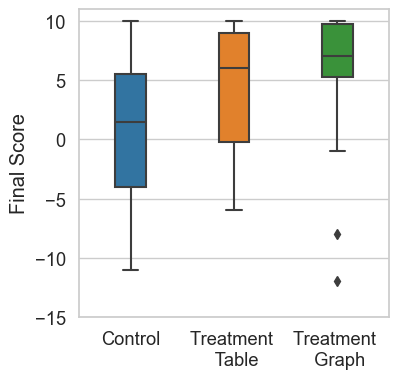}
		\caption{Final Score}
		\label{fig:income-final-score}
	\end{subfigure}
	\begin{subfigure}[b]{0.33\columnwidth}
		\centering
		\includegraphics[width=\linewidth]{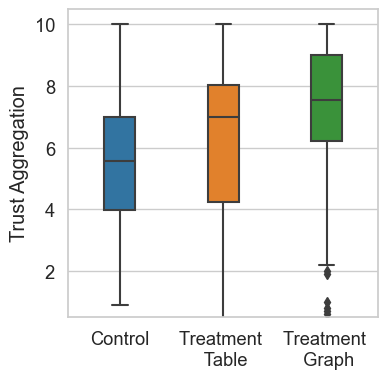}
		\caption{Trust}
		\label{fig:income-trust-stack}
	\end{subfigure}
 	\begin{subfigure}[b]{0.32\columnwidth}
		\centering
		\includegraphics[width=\linewidth]{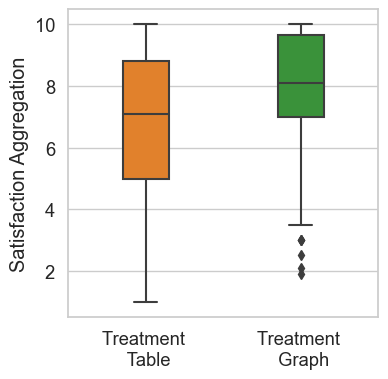} %height=3cm
		\caption{Satisfaction}
		\label{fig:income-explanation-stack}
	\end{subfigure}
	\caption{Domain 1: Income Prediction}
\end{figure}

In this section, we present the results from our experiment for the first domain that used the income prediction dataset. We tested the data normality by using the Shapiro-Wilks test and found that our data was not normally distributed. Therefore, we applied the Mann–Whitney U test, which is a non-parametric test equivalent to the independent samples t-test to perform pairwise comparisons between two groups.

The task prediction scores are shown in Figure~\ref{fig:income-final-score}. We also conduct the Mann-Whitney U test to find whether there is significant difference between two groups. The test shows that participants in both Treatment groups (\textit{TT} and \textit{TG}) performed significantly better than those in the Control group ($p = 0.005 < 0.05, r = 0.41$ and $p < 0.001, r=0.62$ where $r$ is the effect size calculated based on rank correlation). Therefore, the results show that counterfactual explanations of confidence score helps users better \textbf{understand} the AI model's confidence. \textbf{H1a} and \textbf{H1b} are supported. However, there is no significant difference between graph explanations and table explanations in terms of improving users' understanding ($p=0.13>0.05$) (we reject \textbf{H3a}).

Regarding evaluating user trust, Figure~\ref{fig:income-trust-stack} describes the trust scale based on an aggregation of eight metrics (\textit{confident, predictable, reliable, safe, efficient, not wary, perform, decision-making}). We find that both \textit{TT} and \textit{TG} conditions help users better trust the model significantly than \textit{C} condition (C-TT, $p<0.001, r=0.21$; C-TG, $p<0.001, r=0.51$). Thus, \textbf{H2a} and \textbf{H2b} are supported. Moreover, \textit{TG} promotes more trust than \textit{TT} ($p<0.05, r=0.26$) so \textbf{H3b} is supported.

Figure~\ref{fig:income-explanation-stack} shows an aggregation of eight metrics of \textit{Explanation Satisfaction Scale} in both treatment groups (\textit{TT} and \textit{TG}). Users are significantly more satisfied with graph explanations than explanations presented in a table ($p<0.001, r=0.28$). Therefore, \textbf{H3c} is supported.

In summary, giving counterfactual explanations helps users better understand the AI model's confidence score (\textbf{H1a} and \textbf{H1b} are supported). Furthermore, providing counterfactual explanations presented in graphs can improve user trust significantly in the AI model (\textbf{H2a} and \textbf{H2b} are supported). When comparing graph counterfactual explanations and table counterfactual explanations, graph explanations promote user trust and satisfaction more than table explanations (\textbf{H3b} and \textbf{H3c} are supported), however, there is no difference between these two designs in terms of improving understanding (\textbf{H3a} is rejected).
 
\subsection{Results: Domain 2 - HR}

\begin{figure}[ht]
    \begin{subfigure}[b]{0.33\columnwidth}
		\centering
		\includegraphics[width=\linewidth]{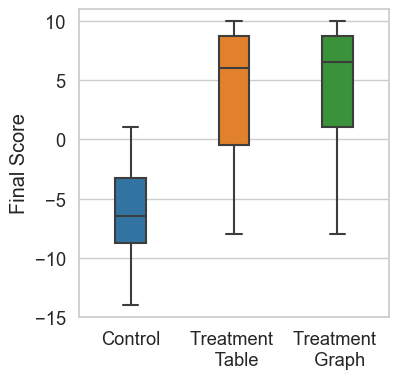} %height=3cm
		\caption{Final Score}
		\label{fig:hr-final-score}
	\end{subfigure}
	\begin{subfigure}[b]{0.33\columnwidth}
		\centering
		\includegraphics[width=\linewidth]{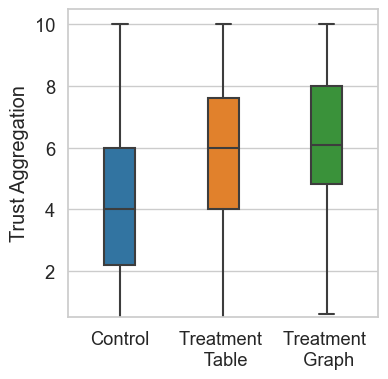}
		\caption{Trust}
		\label{fig:hr-trust-stack}
	\end{subfigure}
 	\begin{subfigure}[b]{0.32\columnwidth}
		\centering
		\includegraphics[width=\linewidth]{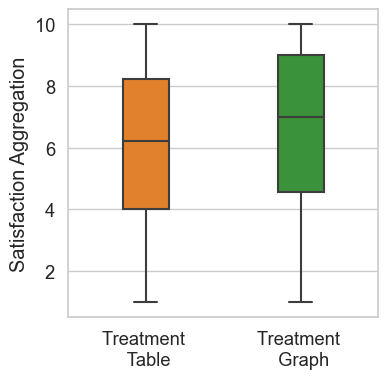} %height=3cm
		\caption{Satisfaction}
		\label{fig:hr-explanation-stack}
	\end{subfigure}
	\caption{Domain 2: HR}
\end{figure}

Since the data in this domain is not normally distributed we also applied Mann–Whitney U test. Figure~\ref{fig:hr-final-score} shows the task prediction scores. Participants from both treatment groups performed significantly better than people in the control group in task prediction (C-TT, $p<0.001,r=0.85$; C-TG, $p<0.001,r=0.87$). Therefore, \textbf{H1a} are \textbf{H1b} are supported. Similar to the first domain, there is no difference ($p=0.86>0.05$) between graph explanations and table explanations (\textbf{H3a} is rejected).

Using Figure~\ref{fig:hr-trust-stack}, graph explanations and table explanations incite more trust than when no explanations given (C-TT, $p<0.001$,$r=0.34$; C-TG, $p<0.001$,$r=0.43$) so \textbf{H2a} and \textbf{H2b} are supported. However, there is no significant difference ($p=0.1>0.05$) between TT and TG in terms of trust (\textbf{H3b} is rejected). Comparing between two forms of counterfactual explanations (TT and TG), Figure~\ref{fig:hr-explanation-stack} shows that there is no difference between these two explanation presentations regarding users' satisfaction ($p=0.06>0.05$). Thus, \textbf{H3c} is rejected.

To sum up, counterfactual explanations of model confidence improve understanding and trust (\textbf{H1a, H1b, H2a} and \textbf{H2b} are supported). This finding is similar to the first domain. However, graph explanations perform no better than table explanations in all three measures \textit{understanding}, \textit{trust} and \textit{satisfaction} (\textbf{H3a}, \textbf{H3b} and \textbf{H3c} are rejected).

\subsection{Results: Summary of Two Domains}

\begin{table}[ht]
	\centering \small
	\setlength{\tabcolsep}{5pt} % Default value: 6pt
	\renewcommand{\arraystretch}{1} % Default value: 1
  \begin{tabular}{lccccccc}
    \toprule
   & \textbf{H1a} & \textbf{H1b} & \textbf{H2a} & \textbf{H2b} & \textbf{H3a} & \textbf{H3b} & \textbf{H3c}\\
  \midrule
\textbf{Domain 1} & $\checkmark$ & $\checkmark$ & $\checkmark$ & $\checkmark$ & $\times$ & $\checkmark$ & $\checkmark$ \\
\textbf{Domain 2} & $\checkmark$ & $\checkmark$ & $\checkmark$ & $\checkmark$ & $\times$ & $\times$ & $\times$ \\
\bottomrule
\end{tabular}
\caption{Summary of hypothesis tests in two domains.  $\checkmark$ represents the hypothesis is supported, $\times$ represents the hypothesis is rejected.}
\label{tab:hypotheses}
\end{table}

Table~\ref{tab:hypotheses} summarises our results of evaluating seven hypotheses. According to the results of two studies, participants who were assigned to treatment groups performed significantly better in task prediction than those in the control group. Similarly, the results show that counterfactual explanations of confidence scores help users trust the AI model more than those who were not given counterfactual explanations. We conclude that \textbf{H1a, H1b, H2a} and \textbf{H2b} are supported in both studies, and therefore, counterfactual explanations of confidence help users better understand the model confidence.

Comparing graph explanations and table explanations, there is no difference in terms of improving users' understanding in the two studies --- \textbf{H3a} is rejected. In study 1, the difference in the task prediction between the two treatment groups is larger than that in study 2. Specifically, effect size in study 1 is $0.23$ ($p=0.13$) and in study 2 is $0.03$ ($p=0.86$). This implies that users in study 1 with graph explanations performed slightly better than those in study 2 with the same condition. Regarding evaluating trust and satisfaction, there are some discrepancies between Domain 1 and 2 when testing \textbf{H3b} and \textbf{H3c}. Based on the effect sizes of H3a in two studies, we argue that since participants in the graph explanations condition in study 1 performed slightly better than those in study 2, they trust and are satisfied more with the graph explanations. We envision the discrepancies of H3b and H3c may be because prior knowledge of participants could affect them doing the tasks in two different domains. Future work could test this idea further.

\section{Conclusion}
This paper proposes two ways to present a counterfactual explanation of model confidence: (1) based on counterfactual examples; and (2) based on counterfactual visualisation. Through two human-subject studies, we show that the counterfactual explanation of model confidence helped users improve their understanding and trust in the AI model. Nonetheless, we found no evidence that counterfactual visualisations are easier to understand. We also found that counterfactual visualisations promote more trust and satisfaction than the counterfactual examples presented in the first domain. However, this conclusion is not true in domain 2 and future work could explore the differences between different counterfactual designs.

In the future, we can improve our model further to generate counterfactual explanations of confidence (or uncertainty) in regression problems when the output is a continuous value instead of a class. Moreover, we can do more studies to explore different presentation options of counterfactual explanations. We also plan to perform more extensive user studies to see whether these explanations can help to improve overall decision making.

\section*{Acknowledgments}
This research was supported by the University of Melbourne Research Scholarship (MRS) and by Australian Research Council (ARC) Discovery Grant DP190103414: Explanation in Artificial Intelligence: A Human-Centred Approach.

\bibliographystyle{named}
\bibliography{references}

\end{document}